%% file: eccv2020submission.tex
\pdfoutput=1
\documentclass[runningheads]{llncs}
\usepackage[dvipdfmx]{graphicx}
\usepackage{comment}
\usepackage{amsmath,amssymb} % define this before the line numbering.
\usepackage{color}
\usepackage{pifont}
\newcommand{\cmark}{\ding{51}}%
\newcommand{\xmark}{\ding{55}}%
\usepackage{here}

\begin{document}
% \linenumbers
\pagestyle{headings}
\mainmatter
\def\ECCVSubNumber{5898}  % Insert your submission number here

\title{Long-Term Human Video Generation of Multiple Futures Using Poses} % Replace with your title

\begin{comment}
\titlerunning{ECCV-20 submission ID \ECCVSubNumber} 
\authorrunning{ECCV-20 submission ID \ECCVSubNumber} 
\author{Anonymous ECCV submission}
\institute{Paper ID \ECCVSubNumber}
\end{comment}
%******************

%\begin{comment}
\titlerunning{Long-Term Human Video Generation of Multiple Futures Using Poses}
\author{Naoya Fushishita\inst{1} \and
Antonio Tejero-de-Pablos\inst{1} \and \\
Yusuke Mukuta\inst{1, 2} \and Tatsuya Harada\inst{1, 2}}
\authorrunning{N. Fushishita et al.}
\institute{The University of Tokyo\\ \email{\{fushishita,antonio-t,mukuta,harada\}@mi.t.u-tokyo.ac.jp} \and
RIKEN}
%\end{comment}
%******************
\maketitle

\begin{abstract}
Generating future video from an input video is a useful task for applications such as content creation and autonomous agents. Especially, prediction of human video is highly important.
While most previous works predict a single future, multiple futures with different behavior can potentially occur. Moreover, if the predicted future is too short (e.g., less than one second), it may not be fully usable by a human or other systems.
In this paper, we propose a novel method for future human pose prediction capable of predicting multiple long-term futures. This makes the predictions more suitable for real applications. After predicting future human motion, we generate future videos based on predicted poses.
First, from an input human video, we generate sequences of future human poses (i.e., the image coordinates of their body-joints) via adversarial learning. Adversarial learning suffers from mode collapse, which makes it difficult to generate a variety of multiple poses. We solve this problem by utilizing two additional inputs to the generator to make the outputs diverse, namely, a latent code (to reflect various behaviors) and an attraction point (to reflect various trajectories). In addition, we generate long-term future human poses using a novel approach based on unidimensional convolutional neural networks. Last, we generate an output video based on the generated poses for visualization. We evaluate the generated future poses and videos using three criteria (i.e., realism, diversity and accuracy), and show that our proposed method outperforms other state-of-the-art works.
\end{abstract}

\input{introduction}

\input{relatedwork}
\input{method}
\input{experiments}
\input{conclusion}

\subsection*{Acknowledgements}
This work was partially supported by JST AIP Acceleration Research Grant Number JPMJCR20U3, and partially supported by the Ministry of Education, Culture, Sports, Science and Technology (MEXT) as “Seminal Issue on Post-K Computer.” We thank Takayuki Hara, Sho Inayoshi, and Kohei Uehara for helpful discussions.

\bibliographystyle{splncs04}
\bibliography{eccv2020submission}

\input{supplemental}
\end{document}

%% file: introduction.tex
%%%%%%%%%%%%%%%%%%%%%%%%%%%%%%%%%%%%%%%%%%%%%%%%%%%%%%%%%%%%%%%%%%%%%%%%%%%%%%%%
%2345678901234567890123456789012345678901234567890123456789012345678901234567890
%        1         2         3         4         5         6         7         8

\section{Introduction}
Future video generation is a very challenging task that has been tackled consistently in the recent years~\cite{lee2018stochastic,mathieu2015deep,villegas2017learning,walker2017pose}, and has applications in different fields (e.g., content creation, autonomous agents, sports analysis). On the one hand, video generation allows for a high-level human interpretability of the predictions. On the other hand, predicting the immediate future from an observed scene is challenging, since several requirements have to be met.
Firstly, since sometimes future is uncertain, there is a range of multiple plausible events that may occur. Thus, future prediction methods that predict a single future \cite{villegas2017learning,butepage2017deep,martinez2017human,gui2018adversarial} may not be versatile enough, since only one possibility of many is considered. Instead, a more realistic setting would involve predicting a variety of plausible futures, as multiple situations can be considered.
Secondly, if the predicted future video is too short, the method would not be realistically usable due to lack of content. For example, if we were to handle a possible dangerous situation predicted in a future video, the predicted time span should be long enough to be able to react in advance. Thus, relatively long predictions are desirable. 
%Furthermore, in case that predictions are interpreted by humans, predictions need to be in an interpretable form, such as natural language sentences or videos. Especially, videos contain a lot of information, and thus, many previous works try to generate them.

% Also, if the prediction is too short, it is hard to interpret, and thus, relatively long predictions are desirable.
% Furthermore, in case that that predictions are interpreted by humans, visualizable predictions are required (i.e., video).

\begin{figure}[tbp]
  \begin{center} %センタリングする
    \includegraphics[width=\linewidth]{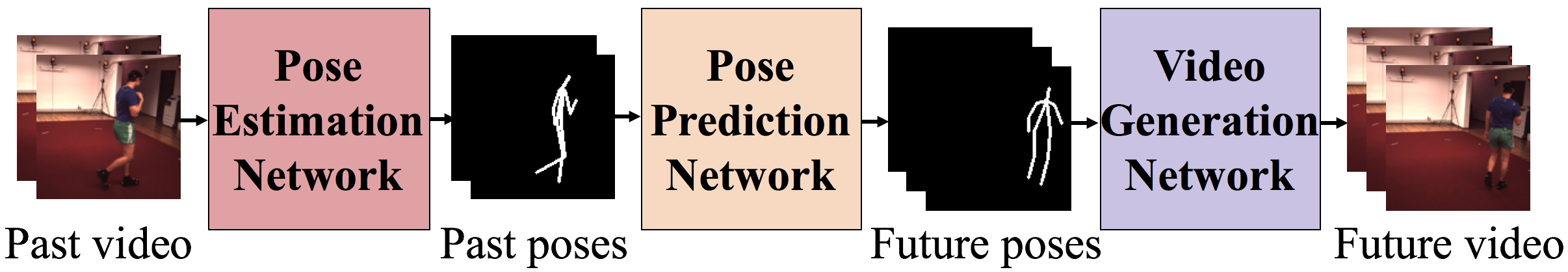}
    \caption{Overview of the proposed method. First, human pose in the input video is estimated. From this, multiple plausible human poses are predicted for a long-term near-future. Finally, the future video based on the predicted poses is generated.} %タイトルをつけ
    \label{fig:pipeline}
    %ラベルをつけ図の参照を可能にする
  \end{center}
\end{figure}

This paper proposes a novel method for video generation of multiple futures from a given input video. Since in many applications (e.g., autonomous agents, sports analysis) prediction of human behaviour is critical, we focus on prediction of human video. As many prior works \cite{villegas2017learning,walker2017pose,yan2017skeleton,cai2018deep} did, we generate future human videos based on  human pose sequences.
First, we estimate the human motion in the input human video. Then, we predict the multiple futures of their movement. Here, the predicted future is long-term (about two to four times longer than short-term future prediction \cite{walker2017pose,mathieu2015deep}).
Finally, after predicting the human behavior, we generate the video representing the predicted future. The overall pipeline is in Figure. \ref{fig:pipeline}.

%Most proficient human video generation methods are based in human pose sequences~.
In order to predict future human motion, many methods model pose sequences by using Recurrent Neural Networks (RNNs) \cite{fragkiadaki2015recurrent,martinez2017human,gui2018adversarial,villegas2017learning,walker2017pose}.
%However, when predicting long-term poses, training an RNN takes a considerably long time because RNNs training cannot be parallelized. 
However, RNNs suffer from the problem of vanishing gradients and error accumulations, which hamper the learning of long data sequences. So, for long-term pose generation, we use unidimensional convolutional neural networks (1D CNNs) instead of RNNs.
We generate predictions of plausible future poses via generative adversarial learning. However, adversarial learning suffers from mode collapse, in which only a few or a single data are generated. We introduce a latent code \cite{chen2016infogan} representing different actions to be able to generate multiple poses. Also, we include a location condition on the generated poses, so human motion is attracted towards different points of the image.

Our contributions are as follows.
\begin{itemize}
   \item We propose a novel method for future human video prediction. We predict multiple futures by (1) imposing a condition to generate various types of motions, and (2) imposing a condition to generate motions towards various locations in the image.
  \item In order to handle long-term future prediction of human behavior, we propose a novel approach for generating human pose sequences using unidimensional convolutional networks.
  \item We provide extensive evaluation of the proposed method to validate our results, and a comparison with state-of-the-art works.
\end{itemize}

%% file: relatedwork.tex
%%%%%%%%%%%%%%%%%%%%%%%%%%%%%%%%%%%%%%%%%%%%%%%%%%%%%%%%%%%%%%%%%%%%%%%%%%%%%%%%
%2345678901234567890123456789012345678901234567890123456789012345678901234567890
%        1         2         3         4         5         6         7         8

\section{Related Work}
%First, Sec. \ref{subsec:gan} explains Generative Adversarial Networks (GAN) \cite{goodfellow2014generative}, which are used for both predicting feasible human behavior and generating the corresponding future videos. Then, Sec. \ref{subsec:video} and Sec. \ref{sec:posepred} introduce related works for video generation tasks and for human pose prediction tasks, respectively. Lastly, Sec. \ref{subsec:video_pose} discusses recent works in human video generation from a human pose input.
%For both, predicting feasible human behavior and generating the corresponding future videos, we use Generative Adversarial Networks (GAN) \cite{goodfellow2014generative}.
\subsection{Generative adversarial networks}
\label{subsec:gan}
Generative Adversarial Networks (GAN) \cite{goodfellow2014generative} is a generative model in which a discriminator is trained to classify between fake data produced by a generator or real data, while the generator is trained to fool the discriminator.
While the output of GAN is generated from latent noise and cannot be controlled, Conditional GAN (CGAN) \cite{mirza2014conditional} includes an input condition like class label that conditions the generated data. InfoGAN \cite{chen2016infogan} unsupervisedly models the relationships between the latent code and the generated images by maximizing the mutual information between them. This allows to apply variations to the generated images without requiring an input label. However, these networks suffer from mode collapse, that is, the model ends up generating only a single or a few predominant data.

\subsection{Automatic video generation}
\label{subsec:video}
GANs are also used for video generation tasks.
Vondrick et al. \cite{vondrick2016generating} proposed VGAN, which generates a foreground video, a background image, and a mask video to merge them. In order to improve coherence in motion and appearance, Ohnishi et al. \cite{ohnishi2018hierarchical} proposed Flow and Texture GAN, which generates optical flow first and then the appearance of the video in a hierarchical architecture.
%For future video generation, Lee et al.~\cite{lee2018stochastic} applied/\edit{proposed?} VAEGAN \cite{larsen2015autoencoding} to
Instead of generating a video from a random latent noise, Mathieu et al. \cite{mathieu2015deep} approached the task of generating a video as a continuation of a video input as a condition. Later, Lee et al. \cite{lee2018stochastic} generated multiple future videos from the same input video.
%\edit{How are these related to the proposed method? (weak points of VAEGAN?)}

\subsection{Human pose prediction}
\label{sec:posepred}
Human pose prediction aims to generate plausible future human behavior from a human behavior input such as coordinates or angles of human joints.
Although many prior works approached this task \cite{brand2000style,wang2008gaussian,taylor2007modeling}, recent developments in deep learning provided an improvement in the results.
Fragkiadaki et al. \cite{fragkiadaki2015recurrent} proposed the Encoder-Recurrent-Decoder model to predict future human poses, which consists of a long short-term memory (LSTM, a kind of RNN) \cite{hochreiter1997long}, an encoder and a decoder.
Similarly, B\"{u}tepage et al. \cite{butepage2017deep} predicted future human poses using an autoencoder-like model.
Gui et al. \cite{gui2018adversarial} proposed a method for future human pose prediction based on adversarial networks with a gated recurrent unit (GRU, a kind of RNN) \cite{cho2014learning}.
While many previous works employ RNNs, these suffer from the vanishing gradients problem: the longer the path between two elements, the worse forward and backward signal propagation \cite{vaswani2015attention,kalchbrenner2016neural}. Also, small errors in the output of the RNN are propagated, and accumulated when generating long sequences. This makes them unsuitable for learning long-term pose sequences.

\subsection{Future video generation using human pose}
\label{subsec:video_pose}

One of the most successful approaches for generating human video is by using a human pose input.
Yan et al. \cite{yan2017skeleton} generated future video from an input frame and a given sequence of future human poses.
%Yan proposed the networks like pix2pix \cite{isola2017image} to generate future image from a given pair of a past image of human and human pose.
Villegas et al. \cite{villegas2017learning} first predicted future human poses as body-joint coordinates using an LSTM and then generated video frame by frame based on generated poses. This approach succeeded in generating long-term videos, but cannot generate multiple futures because the output of an LSTM does not vary for the same input.
Cai et al. \cite{cai2018deep} proposed an adversarial network that generates human pose sequences from latent noise and an action class label, and a network that generates video from the generated poses. This model can be extended to generate a future pose sequence given a past pose sequence, but cannot generate a variety of multiple futures. Also, using an action class label is unsuitable for future prediction, since the action class of the input movement is not available.
Walker et al. \cite{walker2017pose} combined an LSTM and a variational autoencoder (VAE)~\cite{kingma2013auto} to generate multiple human poses from a pose sequence input, and then generate a video using 3D convolutional neural networks. The VAE allows generating multiple human poses, which are then fed to the LSTM to predict a sequence of future poses. However, this approach is unsuitable for long-term future prediction because errors in the LSTM will be accumulated exponentially.

In this paper, we propose a method for long-term video prediction of multiple futures. In order to generate long-term near-future sequences, we leverage unidimensional convolutional neural networks, which allow generating sequences without suffering from the vanishing gradients and error propagation problems. Then, we encourage our network to generate of a variety of multiple futures by using two conditions; a latent code that induces a type of motion, and an attraction point that induces motion towards a location in the image.

%% file: method.tex
%%%%%%%%%%%%%%%%%%%%%%%%%%%%%%%%%%%%%%%%%%%%%%%%%%%%%%%%%%%%%%%%%%%%%%%%%%%%%%%%
%2345678901234567890123456789012345678901234567890123456789012345678901234567890
%        1         2         3         4         5         6         7         8

\section{Methodology}
Figure. \ref{fig:pipeline} shows an overview of our method, which consists of three networks that are trained independently. First, our pose estimation network provides the human pose in a given input video. Then, our pose prediction network generates future human pose sequences that are smooth, varied and long. Finally, our video generation network generates future video corresponding to the generated poses. Since the predicted human poses have a comparatively long duration, and represent a variety of multiple futures, the videos generated using the predicted poses show the same characteristics.

\subsection{Pose estimation network\label{pipeline_openpose}}

%\ref{fig:openpose} shows an overview of our pose estimation network, which
Our pose estimation network estimates the position of the body joints of the human in the video in image coordinates ($xy$ coordinates). Several networks have been proposed in the past \cite{cao2016realtime,newell2016stacked}; we use OpenPose \cite{cao2016realtime}, which has been widely utilized in a variety of related applications.
However, OpenPose is applied to each individual frame and sometimes provides inaccurate estimations (e.g., missing joints). In order to correct this, we leverage sequential information and introduce an autoencoder-like network that takes the entire pose sequence estimated by OpenPose as input. This network consists of an encoder and a decoder, which consist of two fully connected layers each, and calibrates the input joints to be natural as a sequence.
% However, Openpose sometimes provides wrong estimations (e.g., missing joints). Thus, we introduce an autoencoder-like network to correct noisy coordinates given a pose sequence estimated by OpenPose. 
We use the OpenPose network pretrained with the COCO 2016 keypoints challenge dataset \cite{lin2014microsoft}. Thus, only the encoder and the decoder are trained using a dataset with annotations of human joint coordinates (see Sec.~\ref{exp:dataset}), by minimizing the mean squared error between estimated coordinates and those of the ground truth.

%事前学習の際の損失関数は以下の様になる.
%%%
\begin{comment}
\begin{equation}
    \label{eq:openpose_autoencoder}
    \mathcal{L} = \|P_{gt} - D(E(P_{op}))\|_2^2
\end{equation}
\end{comment}
%%%

\subsection{Pose prediction network\label{pipeline_joints}}

\begin{figure}[tbp]
  \begin{center} %センタリングする
    \includegraphics[width=0.92\linewidth]{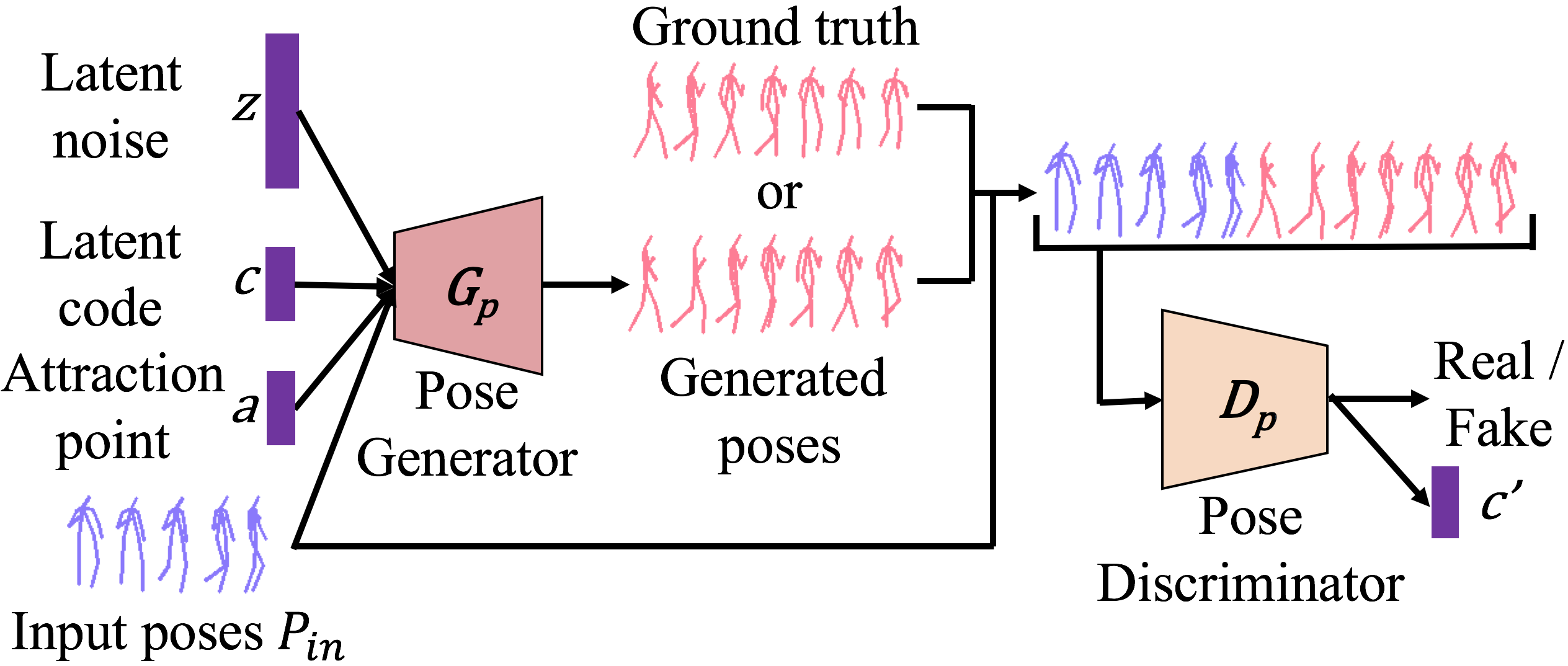}
    \caption{Overview of our pose prediction network. The pose generator ($G_p$) generates a future pose sequence from an input pose sequence given as condition. Then, the pose discriminator ($D_p$) tries to discriminate whether the pose sequence is real (i.e., ground-truth) or fake (i.e., generated). $D_p$ also estimates the latent code $c$ given the generated pose sequence. We use 1D CNN for both $G_p$ and $D_p$. We are able to generate multiple pose sequences by varying the latent code $c$ and the attraction point $a$ randomly.} %タイトルをつけ
    \label{fig:jointgan} %ラベルをつけ図の参照を可能にする
  \end{center}
\end{figure}

Our pose prediction network takes our estimated poses as input and generates future pose sequences. Our generated pose sequences are smoothly connected to the input poses, they have a long-term duration, and represent a variety of multiple futures.

Figure. \ref{fig:jointgan} shows an overview of our pose prediction network. A more detailed figure is available in the Sec. \ref{detailed_pose} of the supplementary material. It consists of two modules: a pose generator ($G_p$) and a pose discriminator ($D_p$).
Let $p_t \in \mathbf{R}^{2N}$ be the human pose at time step $t$. Here, $N$ is the number of joints that compose the pose, and $p_t$ is a vector containing the $xy$ coordinates of $N$ joints at time step $t$. The input of $G_p$ is a latent noise $z$, the input poses from a $T$ frames-long video $P_{in} = (p_0, p_1, ..., p_{T-1})$, a latent code $c \in \mathbf{R}^C$, and an attraction point $a \in \mathbf{R}^2$ ($c$ and $a$ are explained later). The output of $G_p$ is a sequence of $T'$ future human poses $\hat{P}_{gen} = (\hat{p}_T, \hat{p}_{T + 1}, ..., \hat{p}_{T + T' - 1})$ that follow $P_{in}$.

The structure of the network is based on CGAN \cite{mirza2014conditional}; the input poses are included as a condition to $G_p$ and $D_p$.
We use unidimensional convolutional neural networks (1D CNNs) in our generator and discriminator. Although many previous works \cite{fragkiadaki2015recurrent,martinez2017human,gui2018adversarial,villegas2017learning,walker2017pose} used RNNs (i.e., LSTM and GRU) for predicting future human pose sequences, 1D CNNs have advantages over RNNs.
%Whereas RNNs need to be trained sequentially (\ie, one instance at a time), CNNs can be trained in parallel, reducing the execution cost. 
While RNNs output poses one after another, 1D CNNs output an entire pose sequence at once. This frees 1D CNNs from the problem of error accumulation. Furthermore, 1D CNNs can model distant time relationships without being as sensitive as RNNs to the problem of vanishing gradients.
Whereas RNNs need $\mathcal{O}(t)$ steps to predict an element separated $t$ frames from the input, 1D CNNs with a stride width of $s$ need only $\mathcal{O}(\log_s t)$ layers. Since the problem of vanishing gradients gets worse with the number of steps/layers, 1D CNNs seem more suitable to model long-term relationships.
In image generation with a 2D CNN \cite{radford2015unsupervised,karras2019style}, an image is regarded as a three-dimensional entity $\in \mathbf{R}^{H \times W \times 3}$ and convoluted in height and width direction using a two dimensional filter. In our generation task with a 1D CNN, we regard a pose sequence as a two-dimensional entity $P \in \mathbf{R}^{T \times 2N}$ (each row is an individual pose $p$) and convolute it in the height (time) direction with a one-dimensional filter.

CGAN suffers from mode collapse, that is, the generator fails to adequately cover the space of possible predictions and instead generates one or a few prominent modes, ignoring the latent noise. Thus, only modifying the latent noise $z$ is not enough to generate multiple varied pose sequences. To tackle this problem, our method includes two additional inputs to the generator, namely the latent code $c$ and the attraction point $a$. Both are randomly initialized during training, and then used during testing for pose generation from different combinations of $c$ and $a$.
InfoGAN \cite{chen2016infogan} models the relationship between the latent code $c$ and the generated data $G(z, c)$ in an unsupervised way, by maximizing the mutual information between them. Since human actions can be categorized to some extent (e.g., "walking" or "sitting"), we aimed at establishing a correspondence between such action categories and the latent code, and thus, we represent $c$ as a one-hot vector. Note that the pose sequences are not paired with any ground-truth action category label.

The attraction point $a$ represents the $xy$ coordinates of a point in the image space, and is used to train $G_p$ to generate poses constrained to move towards the attraction point. This allows our method to generate multiple varied pose sequences depending on $a$, which in turn is chosen randomly. 
%Such a constraint cannot be applied intuitively in a pixel-based data generation problem (\ie, image or video), since it would be equivalent to constrain the value of a specific pixel to a randomly chosen value. Therefore, while explicitly applying this kind of constraint is not intuitive for pixel-based data generation, it can be used for our joint-coordinates generation task.

\paragraph{Training}
During training, $G_p$ tries to fool the discriminator $D_p$ by generating plausible future pose sequences, while $D_p$ tries to classify whether the pose sequences are real or generated.
The objective function for adversarial learning between $G_p$ and $D_p$ is as follows:
\begin{equation}
    \label{eq:loss_adv_global}
    \begin{split}
    \mathcal{L}_{adv} = &\mathbb{E}_{P_{gt}}[\log D_{p}(P_{gt}|P_{in})] + \mathbb{E}_{z,c,a}[\log (1 - D_{p}(G_p(z,c,a|P_{in})|P_{in})]] + \\
    &\lambda_{gp}\mathbb{E}_{P}[(\|\nabla_{P}D_p(P|P_{in})\|_2 - 1)^2],
    \end{split}
\end{equation}
where $P_{in}$ is the input pose sequence and $P_{gt}$ is ground truth for the predicted pose sequence. We utilize the same gradient penalty as in WGAN-GP \cite{gulrajani2017improved}.

Since it is difficult to directly maximize the mutual information between $c$ and generated data, we introduce an auxiliary probability distribution $Q(c|x)$ and minimize the following function:
\begin{equation}
    \label{eq:loss_info}
    \mathcal{L}_{c} = -\sum_{i=1}^{C}c_{i}\ln{Q(c'|G_p(z,c,a|P_{in}))_{i}}.
\end{equation}
Here, $C$ is the number of categories. As in \cite{chen2016infogan}, $Q$ is implemented by adding two linear layers to the convolutional layers of $D_p$. As depicted in Figure. \ref{fig:jointgan}, $D_p$ also outputs the probability $c'$ of the latent code $c$ given the pose generated by $G_p$.

Our generator $G_p$ is trained to minimize the distance between the generated poses and the attraction point $a$. More concretely, it minimizes the distance between $a$ and the generated coordinate of the waist joint at future frame $t'$: $\hat{p}_{T + t', waist}$. The objective function is:
\begin{equation}
    \label{eq:loss_goal}
    \mathcal{L}_{a} = \frac{1}{T'}\sum_{t'=0}^{T'-1}\|a - \hat{p}_{T + t', waist}\|_2^2.
\end{equation}
In addition, in order to generate smoother pose sequences, we introduce a loss that reduces sudden speed changes between adjacent poses as follows:
\begin{align}
    \label{eq:loss_diff}
    &\mathcal{L}_{diff} = \frac{1}{T'-2}\times \sum_{t'=0}^{T'-3}\|(\hat{p}_{T+t'+2}-\hat{p}_{T+t'+1})-(\hat{p}_{T+t'+1}-\hat{p}_{T+t'})\|_2^2.
\end{align}
In summary, the overall objective function is:
\begin{align}
    \label{eq:loss_all}
    \min_{G_p,Q}\max_{D_p} (\mathcal{L}_{adv} + \lambda_{c}\mathcal{L}_{c} + \lambda_{a}\mathcal{L}_{a} + \lambda_{diff}\mathcal{L}_{diff}),
\end{align}
where $\lambda$s are coefficients to weight the contribution of each loss.

\paragraph{Implementation}
In our implementation, $T$ (the length in frames of the input pose sequences) is 16 and $T'$ (the length in frames of the output pose sequences) is 128. $C$ (the number of categories of the latent code $c$) is 15. These categories correspond to the action classes of the H3.6M dataset \cite{h36m_pami}, which we used for the experiments.
$G_p$ consists of an encoder to encode the input pose sequence and a decoder that generates the predicted pose sequence. The encoder consists of three unidimensional convolutional layers (1D CNN) and the decoder consists of one linear layer and six unidimensional convolutional layers. $D_p$ consists of four unidimensional convolutional layers and, one linear layer to model the $real/fake$ output and two linear layers to model the $c'$ output. We show the details of these network architectures in the Sec. \ref{detailed_pose} of the supplementary material. We set $\lambda_{gp} = 10$, $\lambda_{c} = 2.5$, $\lambda_{a} = 2.5$ and $\lambda_{diff} = 50$.

\subsection{Video generation network\label{pipeline_img}}

\begin{figure}[tbp]
  \begin{center} %センタリングする
    \includegraphics[width=0.9\linewidth]{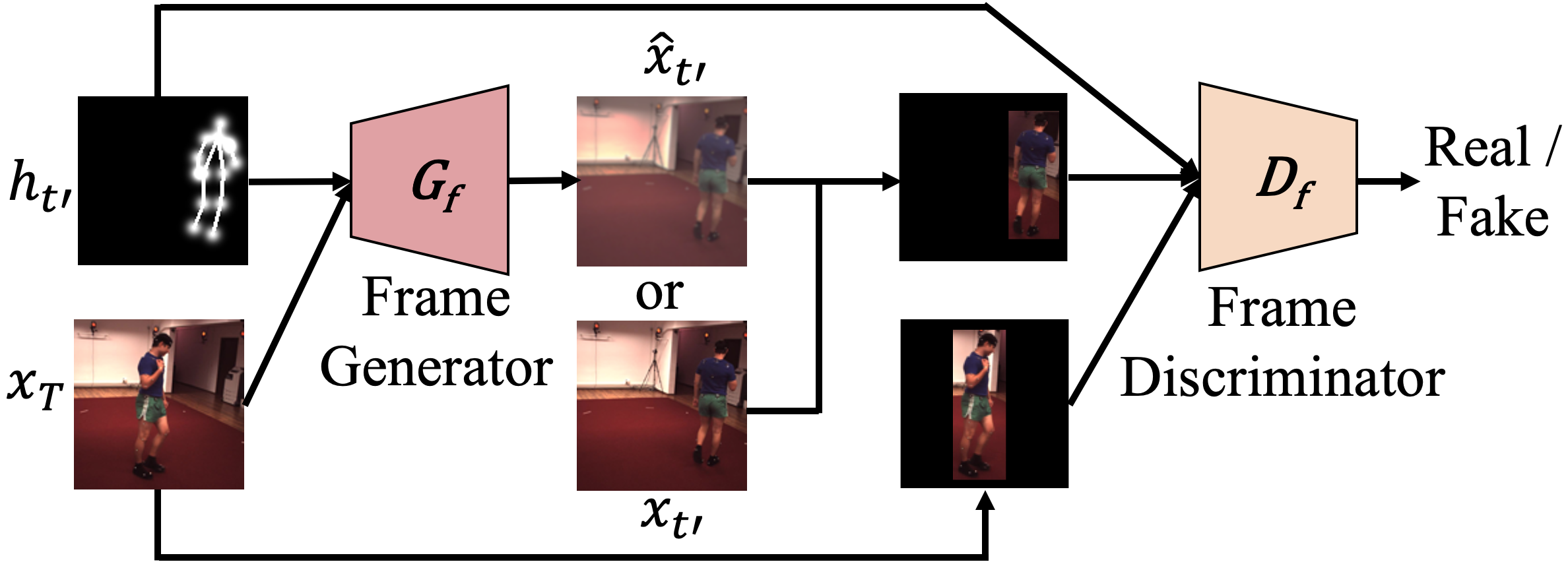}
    \caption{Overview of our video generation network. Each future frame $\hat{x}_{t'}$ is generated from the last frame of the input video $x_T$ and the generated future pose $h_{t'}$. Input images to the discriminator are masked to show only the area around the predicted poses, so that the discriminator can focus on the human.}
    \label{fig:skeleton} %ラベルをつけ図の参照を可能にする
  \end{center}
\end{figure}

Figure. \ref{fig:skeleton} shows an overview of our video generation network. Our video generation network generates future frames with respect to a past frame and our predicted future pose sequence, following an adversarial approach.
We use an architecture based on~\cite{yan2017skeleton}. It generates a single future frame from two inputs, namely the last frame of the input video $x_T$ and a generated future pose $p_{t'}$. The final video is obtained by repeating this for all $T'$ future frames. Not generating the whole video directly at once \cite{vondrick2016generating,walker2017pose} but generating each frame individually \cite{villegas2017learning,yan2017skeleton} increases the image quality.

Before being input to the video generation network, the human pose coordinates generated by the pose prediction network $p_{t'}$ are transformed into a different representation $h_{t'}$. $h_{t'}$ consists of $N+1$ channels with the same height and width as the input video frames. It is built by concatenating a heatmap of $N$ channels, in which each channel represents the position of each joint using a Gaussian distribution centered in the $xy$ coordinates generated by $G_p$, and one channel containing a skeleton that joins those joints.
Thus, for a future frame $t'$, the frame generator $G_f$ takes the last input RGB frame $x_T$ and the predicted future pose $h_{t'}$.
Our $G_f$ follows the U-Net architecture \cite{ronneberger2015unet}.
Inputs $x_T$ and $h_{t'}$ are concatenated in the channel direction. $G_f$ encodes the image with $3+N+1$ channels and decodes it into the future frame $\hat{x}_{t'}$.
Then, our frame discriminator $D_f$ takes the input image $x_T$, the heatmap of the generated future pose $h_{t'}$, and either the real future frame ${x}_{t'}$ or the generated future frame $\hat{x}_{t'}$ and discriminates whether future frame is real or fake (i.e., generated). Since generating a realistic human is more difficult than generating the background, $D_f$ should focus on the foreground human. Therefore, we mask $D_f$ input images to show only the area where the human appears, delimited by the outermost joint coordinates (Figure.~\ref{fig:skeleton}).

\paragraph{Training}
When training our video generation network, instead of using the poses generated by the pose prediction network, we use the poses from the ground truth data. We use three kinds of objective functions, as in \cite{yan2017skeleton}. The first one is the mean absolute error between the pixels in the ground truth video and the generated video:
\begin{equation}
    \label{eq:loss_l1_skeleton}
    \mathcal{L}_{L_1} = \frac{1}{M}\|x_{t'} - G_f(x_T,p_{t'})\|_1,
\end{equation}
where $M$ is the total number of pixels in each frame.

The second one is the adversarial loss:
\begin{comment}
\begin{equation}
    \label{eq:loss_adv_skeleton}
    \begin{split}
    \mathcal{L}_{adv} = & \mathbb{E}_{x_{t'},x_T,p_{t'}}[\log D_f(x_{t'}|x_T,p_{t'})] + \\
    & \mathbb{E}_{x_T,p_{t'}}[\log (1 - D_f(G_f(x_T,p_{t'})|x_T,p_{t'}))] + \\
    &\lambda_{gp}\mathbb{E}_{x}[(\|\nabla_{x}D_f(x|x_T,p_{t'})\|_2 - 1)^2].
    \end{split}
\end{equation}
\end{comment}
\begin{equation}
    \label{eq:loss_adv_skeleton}
    \begin{split}
     & \mathcal{L}_{adv} = \lambda_{gp}\mathbb{E}_{x}[(\|\nabla_{x}D_f(x|x_T,p_{t'})\|_2 - 1)^2] + \\ & \mathbb{E}_{x_{t'},x_T,p_{t'}}[\log D_f(x_{t'}|x_T,p_{t'})] + \mathbb{E}_{x_T,p_{t'}}[\log (1 - D_f(G_f(x_T,p_{t'})|x_T,p_{t'}))].
    \end{split}
\end{equation}
We utilize the gradient penalty of WGAN-GP \cite{gulrajani2017improved}.
 
Lastly, the triplet loss \cite{schroff2015facenet} ensures proper continuity among video frames. Triplet loss addresses three images (i.e., an anchor, a positive and a negative) and minimizes the distance between an anchor and a positive and maximizes the distance between an anchor and a negative. In a video, the L2 distance of adjacent frames should be smaller than that of distant frames. Therefore, when the anchor is $\hat{x}_{t'}$, we set $\hat{x}_{t'+1}$ as positive and $\hat{x}_{t'+5}$ as negative. The concrete objective function is:
\begin{equation}
    \label{eq:loss_tri_skeleton}
    \mathcal{L}_{tri} = \frac{1}{M}[\|\hat{x}_{t'} - \hat{x}_{t'+1}\|_2^2 - \|\hat{x}_{t'} - \hat{x}_{t'+5}\|_2^2 + \alpha]_+ ,
\end{equation}
where $\alpha$ is a margin that is enforced between positive and negative pairs.

In summary, the overall objective function is:
\begin{equation}
    \label{eq:loss_all_skeleton}
    \min_{G_f}\max_{D_f} (\mathcal{L}_{L_1} + \lambda_{adv}\mathcal{L}_{adv} +   \lambda_{tri}\mathcal{L}_{tri}),
\end{equation}
where $\lambda$s are coefficients to weight the contribution of each loss.

\paragraph{Implementation}
$G_f$ consists of an encoder and a decoder, which are connected with skip connections. Both the encoder and the decoder consist of eight convolutional layers each. $D_f$ consists of three parallel convolutional layers, which convolute $h_{t'}$, $x_T$, and $x_{t'}$ or $\hat{x}_{t'}$ respectively, followed by four convolutional layers. We show the details of these network architectures in the Sec. \ref{detailed_video} of the supplementary material. We set $\lambda_{gp} = 10$, $\lambda_{adv} = 0.001$ and $\lambda_{tri} = 10$.

%% file: experiments.tex
%%%%%%%%%%%%%%%%%%%%%%%%%%%%%%%%%%%%%%%%%%%%%%%%%%%%%%%%%%%%%%%%%%%%%%%%%%%%%%%%
%2345678901234567890123456789012345678901234567890123456789012345678901234567890
%        1         2         3         4         5         6         7         8

\section{Experiments}
Evaluating generated video is not straightforward, and normally a single metric is insufficient. While video quality should be evaluated, the diversity of the generated futures is also an important criterion in our method. Furthermore, among all the predicted futures, some of them should be similar to the ground truth.
Following the evaluation in \cite{lee2018stochastic}, we evaluate generated poses and videos from three criteria: realism, diversity and accuracy.

\subsection{Dataset \label{exp:dataset}}
We use the Human3.6M \cite{h36m_pami} dataset to train and evaluate our entire pipeline. Videos in this dataset show 11 actors showing different behavior (e.g., \textit{walking, sitting}). All frames are annotated with the real and image coordinates of 32 body joint positions accurately measured via motion capture. We use 720 videos corresponding to subjects 1, 5, 6, 7, 8 and 9 as train data and 120 videos of subject number 11 as test data. 

We preprocess the videos in the following manner. In order to enlarge actors, videos are cropped by using the outermost poses in the entire sequence, and then resized into 128$\times$128 patches. Since Human3.6M videos have a high frame rate, motion between adjacent frames is small. Therefore, we subsample the video uniformly by taking one every four frames.
We apply two kinds of data augmentation. One is horizontal video flipping. The other is padding frames with black pixels, and randomly cropping patches of size 128$\times$128 containing the human. Since our method masks the human of the input image to the discriminator (see Sec.~\ref{pipeline_img}), this augmentation is not harmful for our method.
We use 14 joints out of the 32 provided: \textit{head, neck, right shoulder, right elbow, right wrist, left shoulder, left elbow, left wrist, right waist, right knee, right foot, left waist, left knee and left foot}. In all experiments, an input of 16 frames long is used to generate future videos of 128 frames long generated as a continuation of the input.

\subsection{Comparison with the related work}
To the best of our knowledge, there is no other work on long-term multiple future video generation, so we compare the performance of our method with two state-of-the-art works in future video generation using human poses. One focuses on generating long-term future video, and the other focuses in generating multiple futures. On the one hand,~\cite{villegas2017learning} predicts long-term future poses by using an LSTM and then generates the video frame by frame. This method avoids error propagation in long sequences since the predicted poses are not input back, but is not capable of generating multiple futures. On the other hand,~\cite{walker2017pose} predicts multiple future poses by using an LSTM and a VAE, and then generates the entire video using a 3D CNN. This method does not seem to be suitable for generating long-term future video, since the predicted poses are repeatedly input back to the LSTM, which causes errors to accumulate.

\subsection{Realism of the generated futures}

\begin{figure}[t]
  \begin{center}
    \begin{tabular}{c}
      % 1
      \begin{minipage}{\hsize}
        \begin{center}
          \includegraphics[clip, width=0.95\linewidth]{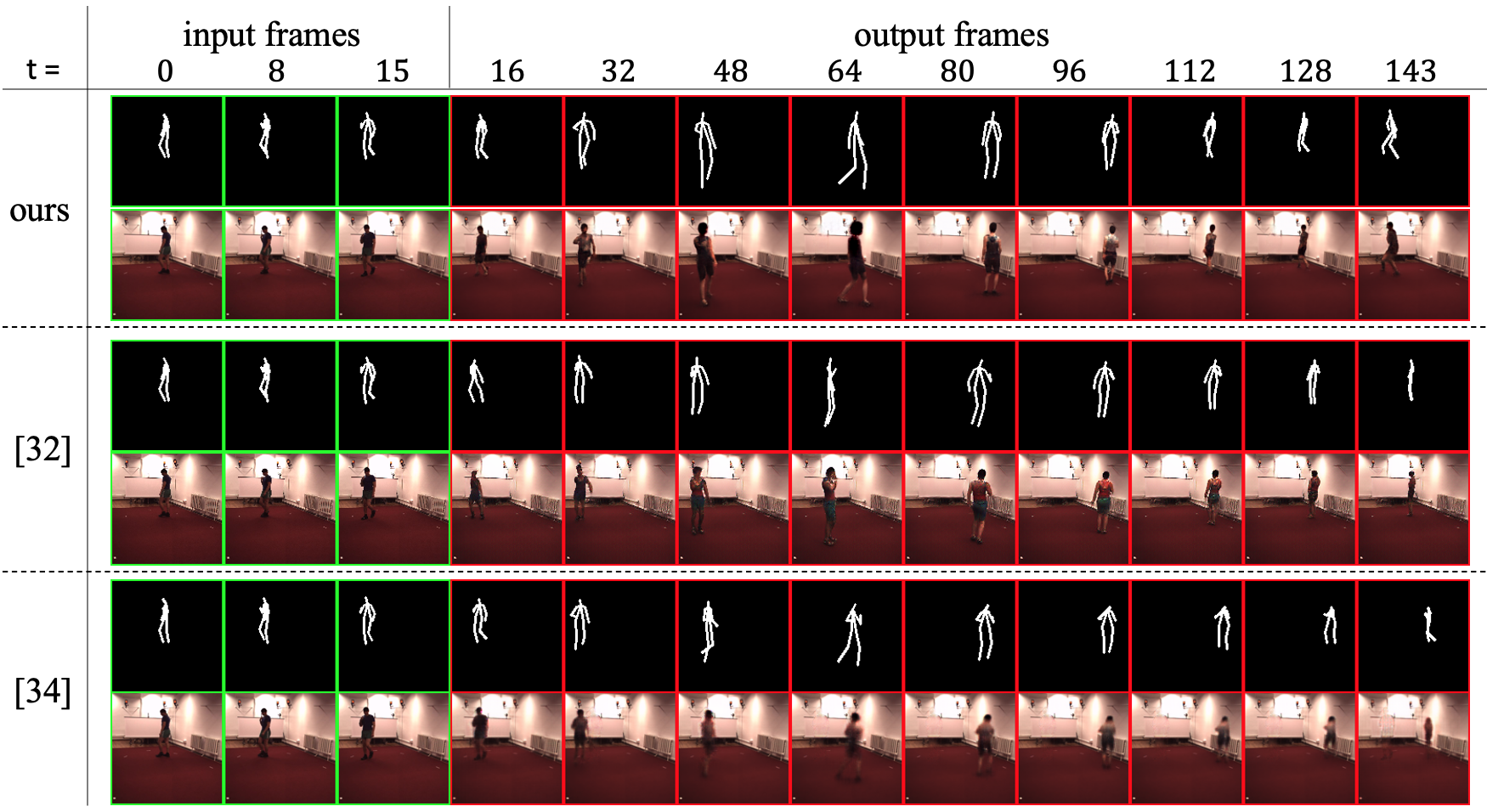} \\
        \end{center}
      \end{minipage}\\

      \begin{minipage}{\hsize}
        \begin{center}
          \includegraphics[clip, width=0.95\linewidth]{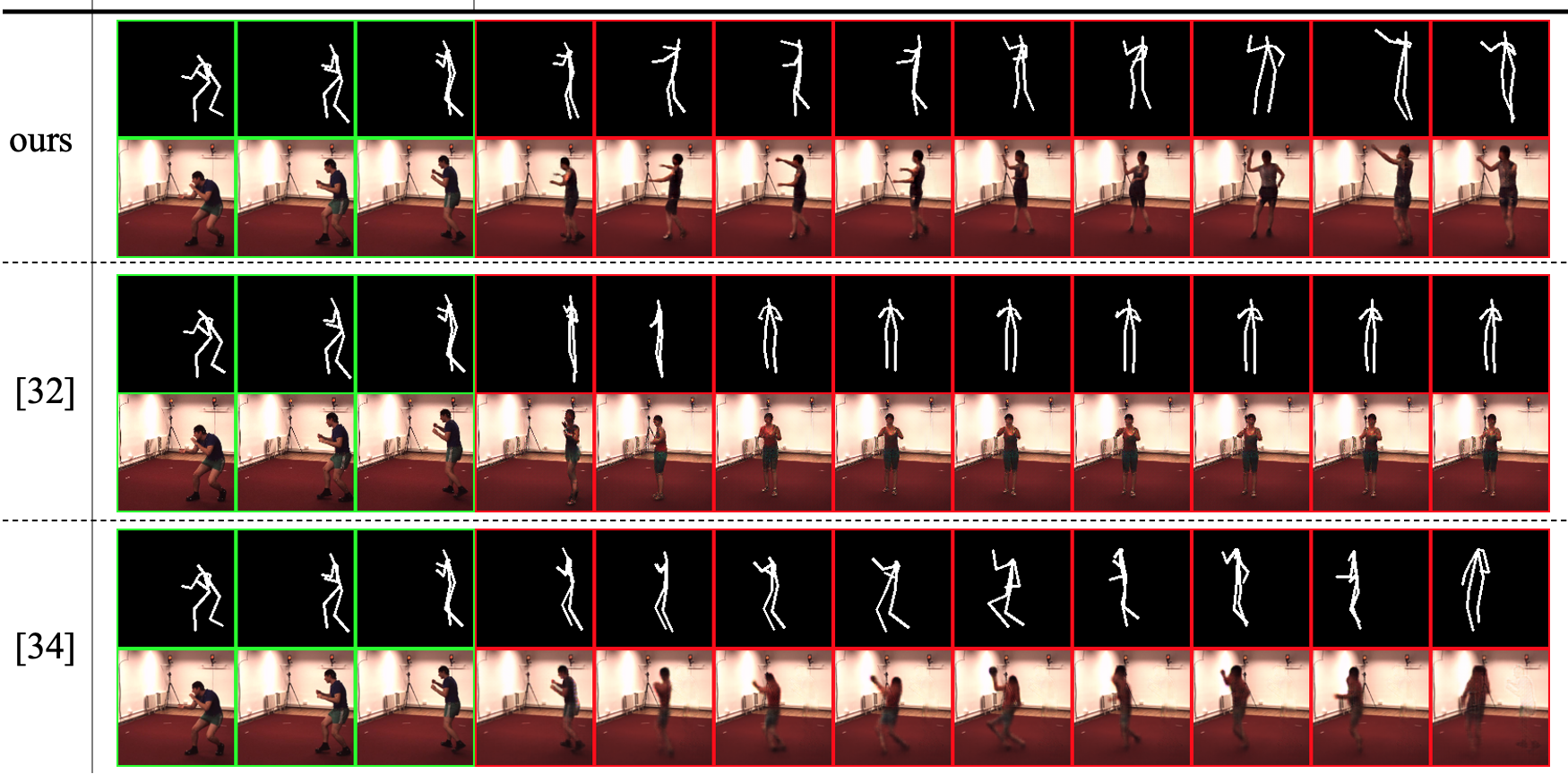} \\
        \end{center}
      \end{minipage}\\
    \end{tabular}
    \caption{Examples of generated poses and frames. Input frames are marked in green and generated frames are marked in red. }
    \label{fig:reality_qual}
  \end{center}
\end{figure}

In this experiment, we evaluate the realism of generated futures via a user study on Amazon Mechanical Turk. We show workers a pair of future poses or a pair of future videos generated by our proposed method and~\cite{villegas2017learning} or~\cite{walker2017pose}, and workers select the one that looks more realistic. 1200 pairs of poses/videos each were evaluated by 120 workers.
\begin{table}[t]
  \centering
  \caption{Evaluation results regarding the realism of our generated futures: Percentage of workers that preferred futures generated by our method vs. those of previous works. Values in brackets represent the p-values of the binomial test.}
  \label{table:amt_result}
  \begin{tabular}{|l||c|c|} \hline
      & vs. \cite{villegas2017learning} & vs. \cite{walker2017pose}\\ \hline \hline
    Pose & 61.3 ($1.98\times10^{-15}$) & 56.2 ($1.08\times10^{-5}$) \\ \hline
    Video & 53.4 ($9.67\times10^{-3}$) & 56.3 ($8.29\times10^{-6}$) \\ \hline
  \end{tabular}
\end{table}

Table \ref{table:amt_result} shows the experimental results. We outperformed both~\cite{villegas2017learning} and~\cite{walker2017pose} in terms of the realism of the generated poses and videos.
Table \ref{table:amt_result} suggests superiority of our pose prediction network, which leverages a unidimensional CNN to predict long-term poses.

Our generated poses were preferred over those from~\cite{villegas2017learning}. As we can see in Figure. \ref{fig:reality_qual}, the pose sequences generated by \cite{villegas2017learning} contain less motion: In the upper example, the person does not move their legs despite they move forward, and in the lower example, the generated pose sequence has almost no movements. Moreover, the connection between input poses and generated poses is not smooth in \cite{villegas2017learning}. On the other hand, the difference between generated videos is smaller. Both, \cite{villegas2017learning} and our method, generate a future frame from the input frame $x_T$ and a future pose $p_{t'}$; however, the larger the difference between $p_{t'}$ and $p_T$ is, the harder generating a realistic frame is. Thus, because the generated poses in \cite{villegas2017learning} are rather motionless (the difference between $p_{t'}$ and $p_T$ is small), they can easily generate future videos with a realistic appearance. Nevertheless, in spite of generating a variety of motions, our videos are preferred for realism.

Also, users preferred our poses and videos to those of~\cite{walker2017pose}. Since \cite{walker2017pose} generates poses one after another using LSTM, errors accumulate and poses tend to gradually deform. Our method does not have such consistency problem because we generate a pose sequence at once via 1D CNN. Also, the videos generated by \cite{walker2017pose} using 3D CNN tend to be blurry compared to those of \cite{villegas2017learning} and ours. This is because we generate the the video frames one by one for each pose, whereas \cite{walker2017pose} generates the video at once.

\subsection{Diversity of the generated futures}\label{sec:div}
We evaluated the diversity of the predicted futures by calculating the distance between futures generated from the same input video as in \cite{lee2018stochastic,zhu2017toward}. The distance becomes larger as the generated futures show more variety. We generate 100 future samples from the same input video, and calculate the distances between all distinct pairs.

We calculate the distance between two future poses as the mean squared error (MSE) of the xy-coordinates of their 14 joints. We use two kind of coordinates systems for this evaluation: One is the absolute coordinates (i.e., image coordinates) and the other is the relative coordinates (i.e., local coordinates with respect to the pose itself). In the relative coordinates, we subtract the coordinates of the right waist joint from all the coordinates. Besides, we calculate the distance between two future videos as the cosine distance of the feature vectors from VGG16 (pretrained by ImageNet \cite{deng2009imagenet}) as in \cite{lee2018stochastic}. This distance is calculated as the average of the five cosine distances between the feature vectors of each of the five pooling layers of VGG16.

Our method is able to generate multiple futures by leveraging a latent code $c$ and an attraction point $a$. To determine the contribution of each one, we did an ablation study evaluating the diversity of the generated videos using four configurations: with $c$ and $a$, with $c$ and without $a$, without $c$ and with $a$, without $c$ and $a$. Also, we compare the results with the multiple futures generated by \cite{walker2017pose}. Since \cite{villegas2017learning} cannot generate diverse futures, it is not included in this evaluation.

\begin{table}[t]
  \centering
  \caption{Evaluation about the diversity of generated poses and videos.}
  \label{table:diversity}
  \begin{tabular}{|l||c|c|c|} \hline
    Method & \begin{tabular}{c}Pose\\(absolute)\\ (MSE)\end{tabular} & \begin{tabular}{c}Pose\\(relative)\\(MSE)\end{tabular} & \begin{tabular}{c}Video\\(Cosine)\end{tabular} \\ \hline \hline
    \cite{walker2017pose} &0.0181& 0.0104 & 0.1447\\ \hline
    w/o $c$, w/o $a$ &0.0102& 0.0062 & 0.2231\\ \hline
    w/o $c$, w/ $a$ &\bf{0.0556}& 0.0162 & 0.3430\\ \hline
    w/ $c$, w/o $a$ &0.0244& 0.0143 & 0.2848\\\hline
    w/ $c$, w/ $a$ &0.0523& \bf{0.0192} & \bf{0.3445}\\\hline
  \end{tabular}
\end{table}

Table \ref{table:diversity} shows the obtained results. Disabling both the latent code $c$ and the attraction point $a$ leads to mode collapse, and the variety worsens. Thus, the average pose distance between two samples in our method is smaller than that of \cite{walker2017pose}. On the other hand, enabling either $c$ or $a$ leads to a greater distance between samples.
Since enabling only $a$ encourages the generated human pose to move to a further location in the image, this configuration leads to the largest distance for poses in absolute coordinates. However, in the case of distances for poses in relative coordinates and distances for videos, enabling both $a$ and $c$ achieves the higher diversity. This proves the efficacy of our attraction point $a$ and latent code $c$ to generate multiple future poses. 

The qualitative results of our evaluation of diversity are included in the Sec. \ref{qualitative_diversity} of the supplementary materials.

\subsection{Accuracy of the generated futures}
\label{subsec:accuracy}
Since this research targets a variety of possible futures, generating futures far from the ground truth is encouraged. However, it would be desirable that at least some futures among the generated are close to the ground truth. Hence, we generate 100 future poses and videos from the same input video, and measure the distance between the ground truth and the sample closest to the ground truth, as in \cite{lee2018stochastic,walker2017pose}. Selecting a ``best'' future is out of the scope of this paper.

Similarity between poses is calculated using the MSE of the image coordinates of the joints (lower is better). Similarity between videos is calculated as the cosine similarity of the feature vectors of VGG 16, and the peak signal-to-noise ratio (PSNR).
We compare the accuracy of the four combinations of adding the latent code $c$ and the attraction point $a$, and the methods in \cite{villegas2017learning} and \cite{walker2017pose}.

\begin{figure*}[t]
  \begin{center}
    \begin{tabular}{c}
      % 1
      \begin{minipage}{\hsize}
        \begin{center}
          \includegraphics[clip, width=\linewidth]{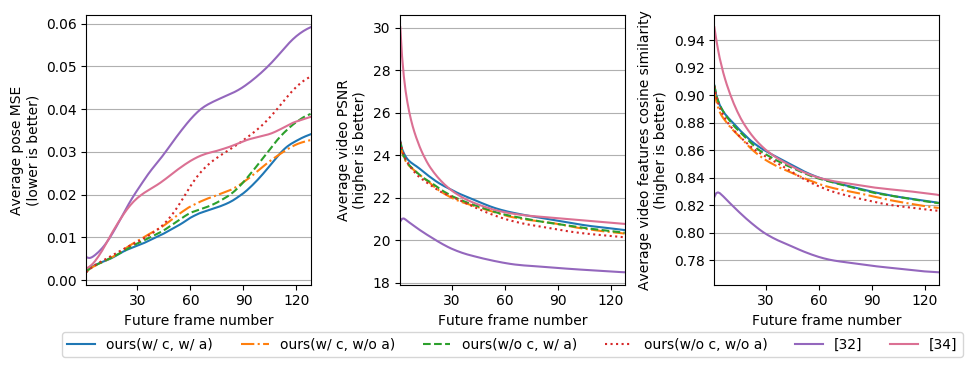}
        \end{center}
      \end{minipage}\\
    \end{tabular}
    \caption{Comparison of the similarity between the generated futures (poses and videos) and the ground truth.}
    \label{fig:accuracy_graph}
  \end{center}
\end{figure*}

Figure. \ref{fig:accuracy_graph} shows the similarity metrics between the ground truth and the pose/video among the generated one hundred with the highest similarity to the ground truth.
With respect to the pose accuracy, using $c$ and $a$ allows for a wider variety of generated poses, thus, there is a higher chance that futures resembling the ground truth are generated. The results suggest that our two additional inputs are effective for not only generating multiple futures but also generating accurate futures.
With respect to the video accuracy, although our method outperforms \cite{walker2017pose} in terms of both accuracy of poses and realism of videos, the video accuracy is slightly lower.
The reason is that \cite{walker2017pose} generates videos that, despite of being blurry, their pixel values are closer to the ground truth. This resembles the phenomenon in which blurred images generated with a pixel-wise loss function (e.g., MSE) tend to have lower MSE with ground truth than sharp images generated using adversarial loss or perceptual loss \cite{blau2018perception,ledig2017photo}. We found that our video accuracy improves by combining our pose prediction network and \cite{walker2017pose}'s video generation network, although the video quality becomes blurry. 

A further analysis is included in the supplementary material (Sec. B.2).

%As referred in \ref{exp:realism}, \cite{walker2017pose} generates video with first few frames of high quality. This is the reason for \cite{walker2017pose} shows high accuracy in first frames. \edit{We don't know why we are outperformed in last few frames.}

%% file: conclusion.tex
%%%%%%%%%%%%%%%%%%%%%%%%%%%%%%%%%%%%%%%%%%%%%%%%%%%%%%%%%%%%%%%%%%%%%%%%%%%%%%%%
%2345678901234567890123456789012345678901234567890123456789012345678901234567890
%        1         2         3         4         5         6         7         8

\section{Conclusions and future work}
In this work, we present a novel method for generating long-term future videos of multiple futures from an input human video using a hierarchical approach: first predicting future human poses and then generating the future video. We propose a novel network to predict long-term future human pose sequences by using unidimensional convolutional neural network in adversarial training. Also, we propose two additional inputs that allow predicting a variety of multiple futures: a latent code and an attraction point. Finally, videos generated with our predicted poses are also long and multiple. Experimental results on the realism, diversity, and accuracy of the generated poses and videos show the superiority of the proposed method over the state-of-the-art.

As our future work, since our method generates videos frame by frame, videos with a higher resolution could be generated by leveraging the latest image generation techniques using GAN \cite{brock2018large,karras2019style}. Also, we plan to tackle the limitations of our method; for example, generating videos with moving background.

%\section{今後の課題と展望}
%本研究で提案した骨格生成ネットワークは,過去の骨格の動きのみから未来の骨格を生成する.そのため,周りの環境などは考慮していないため,例えば本来は壁があるところを歩くような骨格を生成してしまう恐れがある.これを改良するために,骨格生成ネットワークの入力に何らかの形で環境情報を加える必要がある.本研究で用いた動画像生成ネットワークでは,過去の骨格と生成された未来の骨格に大きな差がある場合,対応する未来のフレームの生成に失敗しうることがわかった.より本物に近い動画像を生成するためには,骨格から動画像を生成するためのよりよいネットワークを開発する必要がある.本研究の手法は,人間しか写っている動画像にしか適用することができないという欠点がある.また,人間の動きのみを考慮しているため,人間以外の物体の動きは無視されてしまう.そのため,より一般的なモデルにするためには人間から物体全体への拡張が必要である.また,本研究の手法では物体間の干渉などを考えていない.そのため,例えば人間が椅子をすり抜けるような動画像が生成されてしまう可能性がある.しかし,これでは現実世界のシステムに応用することが難しい.物体間の干渉を扱うために,動画像をそのまま二次元の世界として扱うのではなく,三次元の世界を再構成する必要がある.

%% file: supplemental.tex
\clearpage
\onecolumn
\section*{Supplementary materials}
\setcounter{section}{0}
\renewcommand{\thesection}{\Alph{section}}

\section{Detailed implementations}
\subsection{Pose prediction network}
\label{detailed_pose}
\begin{figure*}[h]
  \begin{center} %センタリングする
    \includegraphics[width=0.90\linewidth]{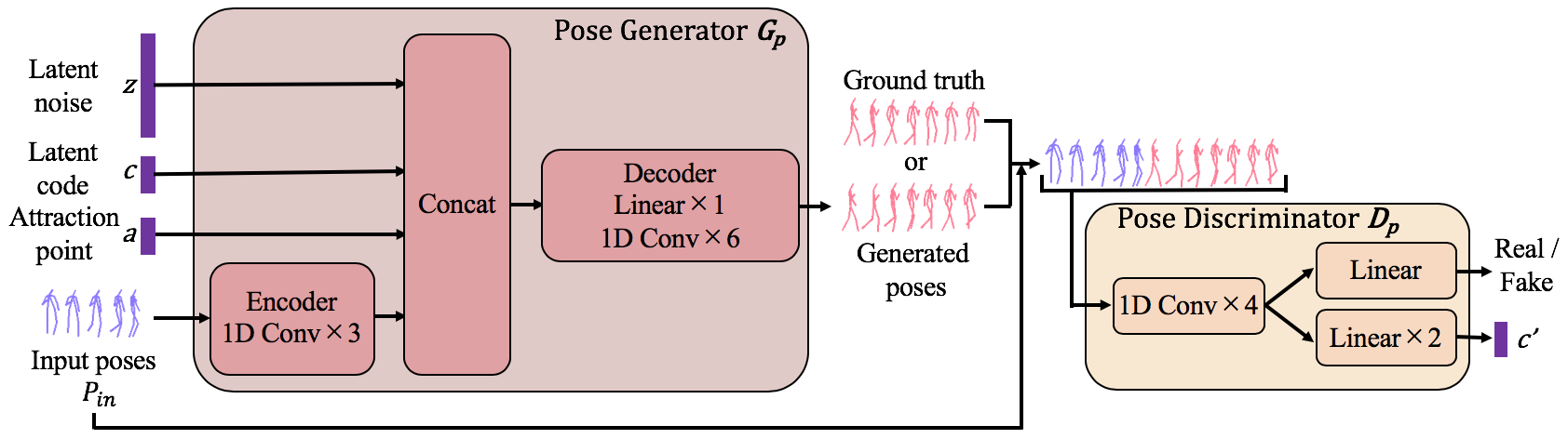}
    \caption{Details of the components of our pose prediction network.}
    \label{fig:pose_prediction_network_detail} %ラベルをつけ図の参照を可能にする
    \vspace{-5mm}
  \end{center}
\end{figure*}

We show the details of the components of our pose prediction network on Figure. \ref{fig:pose_prediction_network_detail} and the details of the architecture and hyper parameters of it on Table \ref{table:architecture_pose_prediction_network}.

\begin{table*}[h]
  \begin{center}
  \caption{Details of the architecture and hyper parameters of our pose prediction network (BN$=$batch normalization).}
  \label{table:architecture_pose_prediction_network}
  \scalebox{0.77}{
    \begin{tabular}{lccccccc} \hline
      Operation & Kernel & Strides & Padding & Input shape & Output shape & BN? & Activation function\\ \hline \hline
    Encoder & & & & (time, channel) & (time, channel) \\ \hline
    1D Conv & 4 & 2 & 1 & (16, 2$N$) & (8, 4$N$) & \xmark & Leaky ReLU \\ 
    1D Conv & 4 & 2 & 1 & (8, 4$N$) & (4, 8$N$) & \cmark & Leaky ReLU \\ 
    1D Conv & 4 & 2 & 1 & (4, 8$N$) & (2, 16$N$) & \cmark & Leaky ReLU \\ \hline
    Decoder \\ \hline
    Linear & & & & 500 + $C$ + 2 + 2$\times$16$N$ & 64$N\times$4 & \cmark & \\
    1D Deconv & 4 & 2 & 1 & (4, 64$N$) & (8, 64$N$) & \cmark & ReLU \\ 
    1D Deconv & 4 & 2 & 1 & (8, 64$N$) & (16, 64$N$) & \cmark & ReLU \\ 
    1D Deconv & 4 & 2 & 1 & (16, 64$N$) & (32, 64$N$) & \cmark & ReLU \\ 
    1D Deconv & 4 & 2 & 1 & (32, 64$N$) & (64, 32$N$) & \cmark & ReLU \\ 
    1D Deconv & 4 & 2 & 1 & (64, 32$N$) & (128, 16$N$) & \cmark & ReLU \\ 
    1D Deconv & 1 & 1 & 0 & (128, 16$N$) & (128, 2$N$) & \xmark & tanh \\ \hline \hline
    Discriminator \\ \hline
    1D Conv & 4 & 2 & 1 & (144, 2$N$) & (72, 4$N$) & \xmark & Leaky ReLU \\ 
    1D Conv & 4 & 2 & 1 & (72, 4$N$) & (36, 8$N$) & \cmark & Leaky ReLU \\ 
    1D Conv & 4 & 2 & 1 & (36, 8$N$) & (18, 16$N$) & \cmark & Leaky ReLU \\ 
    1D Conv & 4 & 2 & 1 & (18, 16$N$) & (9, 32$N$) & \cmark & Leaky ReLU \\ 
    \multicolumn{4}{l}{Linear (for real/fake)} & 9$\times$32$N$ & 1 & \xmark & \\ 
    \multicolumn{4}{l}{Linear (for $c'$)} &  9$\times$32$N$ & 128 & \cmark & Leaky ReLU \\ 
    \multicolumn{4}{l}{Linear (for $c'$)} & 128 & $C$ & \xmark & \\ \hline \hline
    & \\
    \multicolumn{3}{l}{Hyper parameters} \\ \hline
    \multicolumn{2}{l}{Optimizer} & \multicolumn{6}{l}{Adam ($\alpha = 2 \times 10^{-4}, \beta_1 = 0.5, \beta_2 = 0.999$)} \\
    \multicolumn{2}{l}{Batch size} & 32 \\
    \multicolumn{2}{l}{Latent noise dimension} & 500 \\
    \multicolumn{2}{l}{Latent code dimension ($C$)} & 15 \\
    \multicolumn{2}{l}{Number of joints ($N$)} & 14 \\
    \multicolumn{2}{l}{Leaky ReLU slope} & 0.2 \\ \hline
    
  \end{tabular}
  }
  \end{center}
\end{table*}

\clearpage
\subsection{Video Generation Network}
\label{detailed_video}
We show the details of the architecture and hyper parameters of our video generation network on Table \ref{table:architecture_video_generation_network}.

\begin{table}[h]
  \begin{center}
  \caption{Details of the architecture and hyper parameters of our video generation network. Note that the output and input shapes between contiguous layers is different in the decoder since the encoder and the decoder are connected with skip connections (BN$=$batch normalization).}
  \label{table:architecture_video_generation_network}
  \scalebox{0.77}{
  \begin{tabular}{lccccccc} \hline
      Operation & Kernel & Strides & Padding & Input shape & Output shape & BN? & Activation function\\ \hline \hline
    Encoder & & & & (height, width, channel) & \\ \hline
    2D Conv & 3 & 1 & 1 & (128, 128, 3+$N$+1) & (128, 128, 64) & \xmark & Leaky ReLU \\ 
    2D Conv & 4 & 2 & 1 & (128, 128, 64) & (64, 64, 128) & \cmark & Leaky ReLU \\ 
    2D Conv & 4 & 2 & 1 & (64, 64, 128) & (32, 32, 256) & \cmark & Leaky ReLU \\ 
    2D Conv & 4 & 2 & 1 & (32, 32, 256) & (16, 16, 512) & \cmark & Leaky ReLU \\ 
    2D Conv & 4 & 2 & 1 & (16, 16, 512) & (8, 8, 512) & \cmark & Leaky ReLU \\ 
    2D Conv & 4 & 2 & 1 & (8, 8, 512) & (4, 4, 512) & \cmark & Leaky ReLU \\ 
    2D Conv & 4 & 2 & 1 & (4, 4, 512) & (2, 2, 512) & \cmark & Leaky ReLU \\ 
    2D Conv & 4 & 2 & 1 & (2, 2, 512) & (1, 1, 512) & \cmark & Leaky ReLU \\ \hline
    
    Decoder \\ \hline
    2D Deconv & 4 & 2 & 1 & (1, 1, 512) & (2, 2, 512) & \cmark & ReLU \\ 
    Dropout \\
    2D Deconv & 4 & 2 & 1 & (2, 2, 1024) & (4, 4, 512) & \cmark & ReLU \\ 
    Dropout \\
    2D Deconv & 4 & 2 & 1 & (4, 4, 1024) & (8, 8, 512) & \cmark & ReLU \\ 
    Dropout \\
    2D Deconv & 4 & 2 & 1 & (8, 8, 1024) & (16, 16, 512) & \cmark & ReLU \\ 
    2D Deconv & 4 & 2 & 1 & (16, 16, 1024) & (32, 32, 256) & \cmark & ReLU \\ 
    2D Deconv & 4 & 2 & 1 & (32, 32, 512) & (64, 64, 128) & \cmark & ReLU \\ 
    2D Deconv & 4 & 2 & 1 & (64, 64, 256) & (128, 128, 64) & \cmark & ReLU \\ 
    2D Conv & 3 & 1 & 1 & (128, 128, 128) & (128, 128, 3) & \xmark & \\ \hline \hline
    
    Discriminator \\ \hline
    2D Conv (for $x_{t'}$) & 4 & 2 & 1 & (128, 128, 3) & (64, 64, 32) & \xmark & Leaky ReLU \\ 
    2D Conv (for $x_T$) & 4 & 2 & 1 & (128, 128, 3) & (64, 64, 32) & \xmark & Leaky ReLU \\ 
    2D Conv (for $h_{t'}$) & 4 & 2 & 1 & (128, 128, $N$+1) & (64, 64, 32) & \xmark & Leaky ReLU \\
    2D Conv & 4 & 2 & 1 & (64, 64, 96) & (32, 32, 128) & \cmark & Leaky ReLU \\ 
    2D Conv & 4 & 2 & 1 & (32, 32, 128) & (16, 16, 256) & \cmark & Leaky ReLU \\ 
    2D Conv & 4 & 2 & 1 & (16, 16, 256) & (8, 8, 512) & \cmark & Leaky ReLU \\ 
    2D Conv & 3 & 1 & 1 & (8, 8, 512) & (8, 8, 1) & \xmark &  \\ \hline \hline
    & \\
    \multicolumn{3}{l}{Hyper parameters} \\ \hline
    \multicolumn{2}{l}{Optimizer} & \multicolumn{6}{l}{Adam ($\alpha = 2 \times 10^{-4}, \beta_1 = 0.5, \beta_2 = 0.999$)} \\
    \multicolumn{2}{l}{Batch size} & 10 \\
    \multicolumn{2}{l}{Number of joints ($N$)} & 14 \\
    \multicolumn{2}{l}{Leaky ReLU slope} & 0.2 \\
    \multicolumn{2}{l}{Dropout rate} & 0.5 \\ \hline
    
  \end{tabular}
  }
  \end{center}
\end{table}

\clearpage
\section{Additional results}
\subsection{Qualitative results of diversity}
\label{qualitative_diversity}
\begin{figure*}[h]
  \begin{center}
    \begin{tabular}{c}
      % 1
      \begin{minipage}{\hsize}
        \begin{center}
          \includegraphics[clip, width=\linewidth]{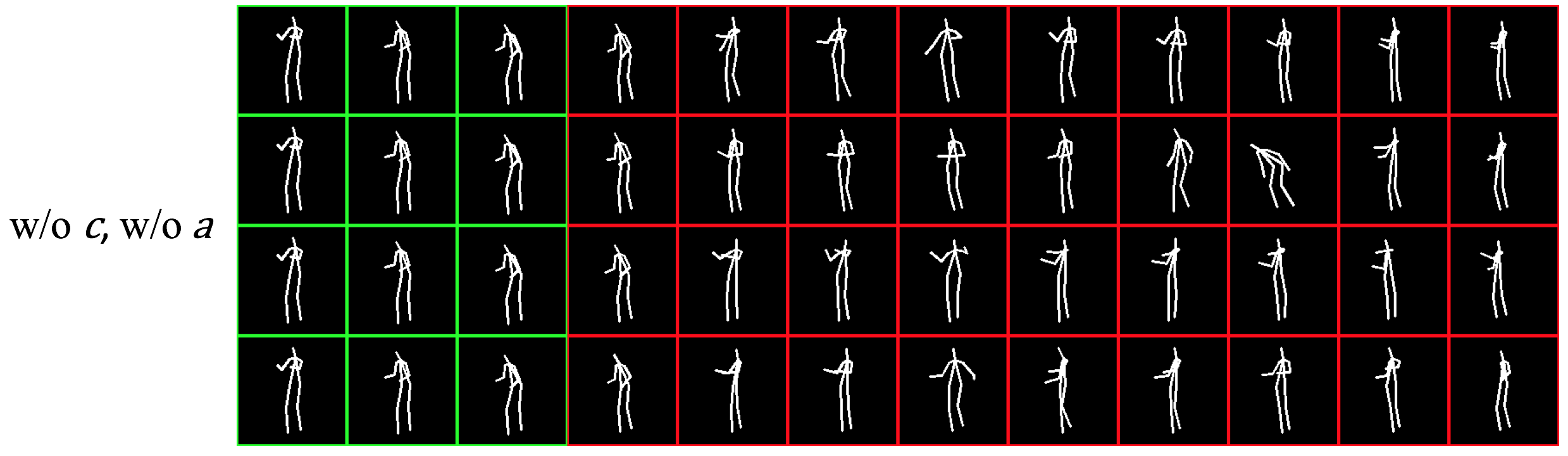} \\
        \end{center}
      \end{minipage}\\

      \begin{minipage}{\hsize}
        \begin{center}
          \includegraphics[clip, width=\linewidth]{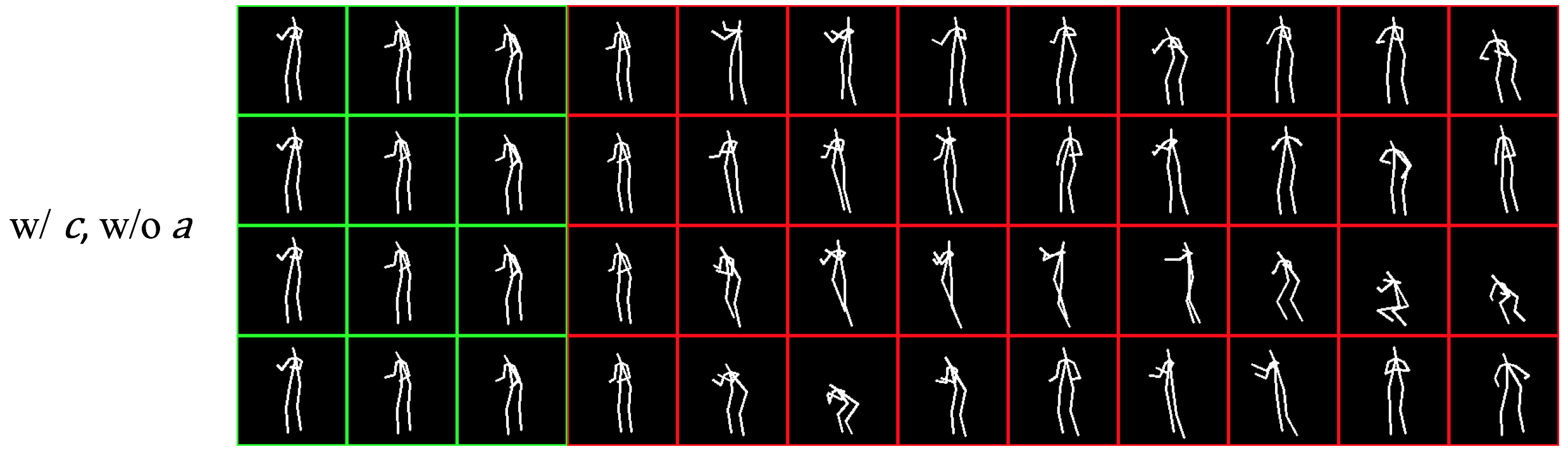} \\
        \end{center}
      \end{minipage}\\
      
      \begin{minipage}{\hsize}
        \begin{center}
          \includegraphics[clip, width=\linewidth]{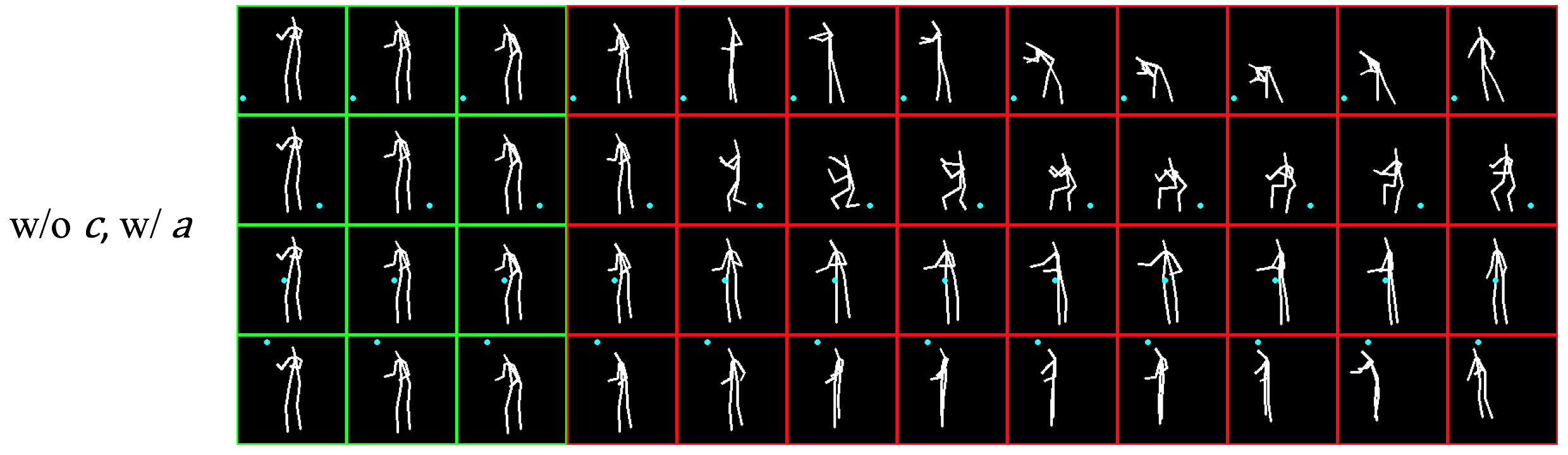} \\
        \end{center}
      \end{minipage}\\
      
      \begin{minipage}{\hsize}
        \begin{center}
          \includegraphics[clip, width=\linewidth]{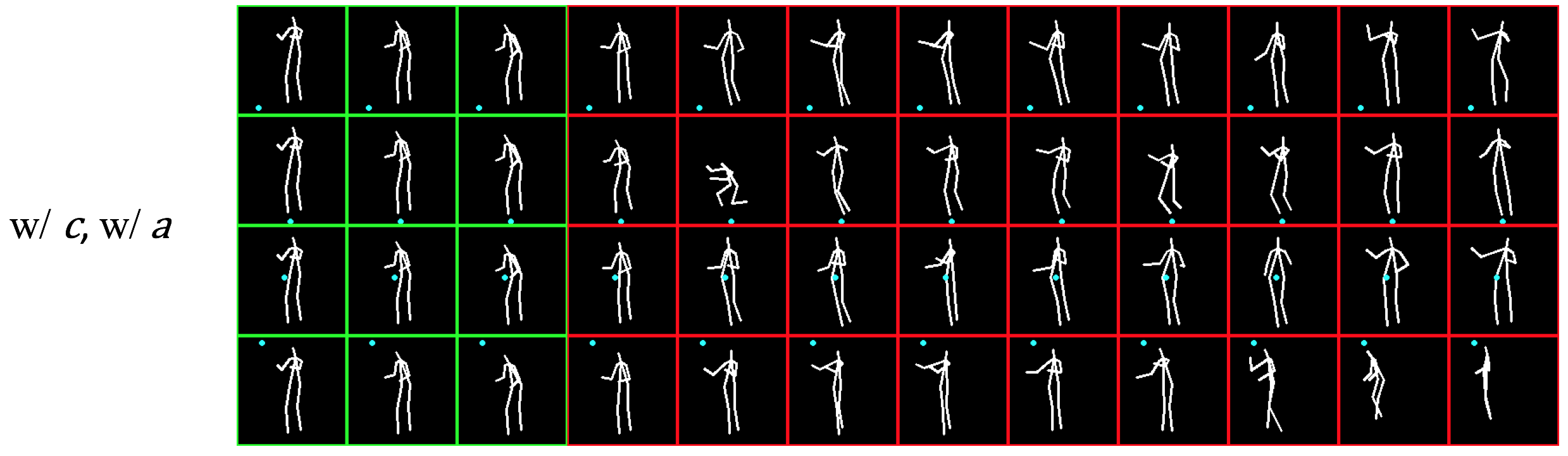} \\
        \end{center}
      \end{minipage}\\
    \end{tabular}
    \caption{Examples of multiple poses predicted from the same input. Input frames are marked in green and generated frames are marked in red. The light blue dots in the bottom two figures represent the attraction points $a$ which are initialized randomly. By choosing a different $a$ and $c$, the predictions of our method also vary.}
    \label{fig:diversity_qual}
  \end{center}
\end{figure*}

Figure. \ref{fig:diversity_qual} shows examples of multiple pose sequences generated from the same input by our method. When neither the latent code $c$ nor the attraction point $a$ are enabled, mode collapse occurs and the generated poses show less diversity. 

When we utilize only the latent code $c$, the generated poses become more varied. Although we set the dimension of the latent code equal to the number of action categories in Human3.6M \cite{h36m_pami} (i.e., 15) and we aimed at establishing a correspondence between the latent code and the action categories, the latent code does not seem to acquire such correspondence. We think this is because of two reasons. The first reason is that the action categories in Human3.6M are not separated completely. For example, although Human3.6M contains a "sitting" class, the "eating" class also shows the sitting behavior. This results in the difference between the number of action categories in the dataset and the actual number of behaviors, and makes the dimension we set not suitable. The second reason is that some action categories rarely occur right after certain action categories (e.g., when an input pose sequence shows a walking person, they rarely lie down within the next few seconds). This makes it difficult for the latent code to cover all the action categories. As a result, rather than covering a perfect match of all action categories, our latent code is able to capture low-level motion differences (e.g., the movement of the arms and the timing to start moving). Since our goal is not to control the generated motion but to predict multiple plausible future motions, we do not regard it as a problem.

The light blue dots in the lower images of Figure. \ref{fig:diversity_qual} represent the attraction points, which are initialized randomly. When we utilize only the attraction point $a$ (we disable $c$), the person in the image gets close to it. However, although generating motions that approach the point linearly is the optimal way to minimize the attraction loss (eq. \ref{eq:loss_goal}), sometimes the person does not reach the point completely. This is because there is a trade-off between the attraction point loss and the other losses.

When we enable both the attraction point $a$ and the latent code $c$, the effect of the attraction point $a$ becomes smaller (e.g., in the top row of the bottom figure of Figure. \ref{fig:diversity_qual}, the person does not sit or lie while the attraction point is under the person). This is the reason why the \textit{absolute} pose diversity is lower than the diversity when only the attraction point is enabled in Table \ref{table:diversity}. However, by combining the attraction point and the latent code, the \textit{relative} pose diversity increases.

\subsection{Further analysis about accuracy}

\begin{figure*}[b]
  \begin{center}
    \begin{tabular}{c}
      % 1
      \begin{minipage}{\hsize}
        \begin{center}
          \includegraphics[clip, width=\linewidth]{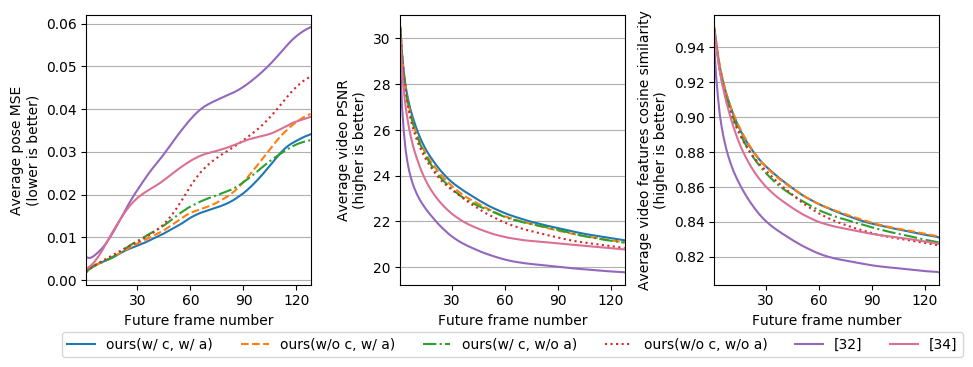}
        \end{center}
      \end{minipage}\\
    \end{tabular}
    \caption{Comparison of the similarity between the generated futures (poses and videos) and the ground truth. In all the curves, \cite{walker2017pose}'s video generation network is used.}
    \label{fig:accuracy_graph_posevae}
  \end{center}
\end{figure*}

\begin{figure*}[b]
  \begin{center}
    \begin{tabular}{c}
      % 1
      \begin{minipage}{\hsize}
        \begin{center}
          \includegraphics[clip, width=\linewidth]{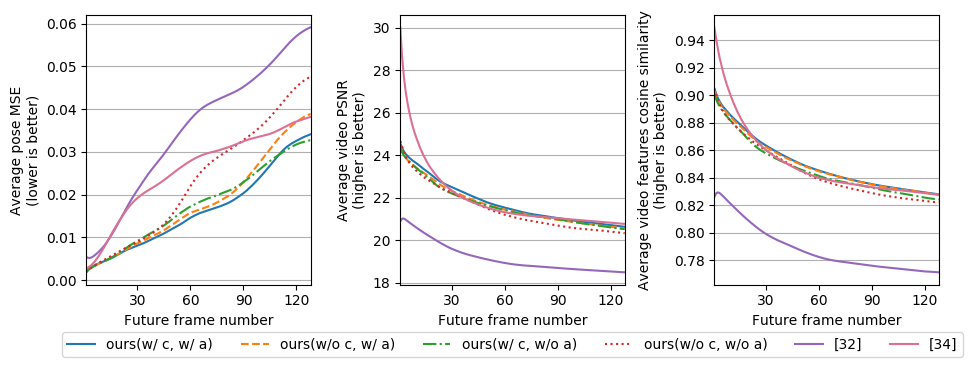}
        \end{center}
      \end{minipage}\\
    \end{tabular}
    \caption{Comparison of the similarity between the generated futures (poses and videos) and the ground truth. In the curves of our method, the adversarial loss in the video generation network $\lambda_{adv}$ is set to 0.}
    \label{fig:accuracy_graph_adv0}
  \end{center}
\end{figure*}

As we remark in Sec. \ref{subsec:accuracy}, by combining our pose prediction network and \cite{walker2017pose}'s video generation network, video accuracy increases at the sacrifice of the video quality (blurriness increases). Figure. \ref{fig:accuracy_graph_posevae} shows the accuracy results when using \cite{walker2017pose}'s video generation network in all the curves. When using the same video generation network, the accuracy of the videos becomes almost equal to the accuracy of the poses and the configurations of our method outperform the related work. Also, in our video generation network, we utilize an adversarial loss (eq. \ref{eq:loss_adv_skeleton}) to sharpen the generated videos. By disabling this loss (i.e., setting $\lambda_{adv} = 0$), our videos become a little blurry but the accuracy improves, becoming comparable to the video accuracy of \cite{walker2017pose} (Figure. \ref{fig:accuracy_graph_adv0}). From this fact, we can conclude the existence of a trade-off relationship between the video quality and the video accuracy.

\clearpage

%% file: eccv2020submission.bbl
\begin{thebibliography}{10}
\providecommand{\url}[1]{\texttt{#1}}
\providecommand{\urlprefix}{URL }
\providecommand{\doi}[1]{https://doi.org/#1}

\bibitem{blau2018perception}
Blau, Y., Michaeli, T.: The perception-distortion tradeoff. In: Computer Vision
  and Pattern Recognition (2018)

\bibitem{brand2000style}
Brand, M., Hertzmann, A.: Style machines. In: ACM International Conference on
  Computer Graphics and Interactive Techniques (2000)

\bibitem{brock2018large}
Brock, A., Donahue, J., Simonyan, K.: Large scale gan training for high
  fidelity natural image synthesis. In: International Conference on Learning
  Representations (2019)

\bibitem{butepage2017deep}
B{\"u}tepage, J., Black, M.J., Kragic, D., Kjellstr{\"o}m, H.: Deep
  representation learning for human motion prediction and classification. In:
  Computer Vision and Pattern Recognition (2017)

\bibitem{cai2018deep}
Cai, H., Bai, C., Tai, Y.W., Tang, C.K.: Deep video generation, prediction, and
  completion of human action sequences. In: European Conference on Computer
  Vision (2018)

\bibitem{cao2016realtime}
Cao, Z., Simon, T., Wei, S.E., Sheikh, Y.: Realtime multi-person 2d pose
  estimation using part affinity fields. In: Computer Vision and Pattern
  Recognition (2017)

\bibitem{chen2016infogan}
Chen, X., Duan, Y., Houthooft, R., Schulman, J., Sutskever, I., Abbeel, P.:
  Infogan: Interpretable representation learning by information maximizing
  generative adversarial nets. In: Neural Information Processing Systems (2016)

\bibitem{cho2014learning}
Cho, K., van Merrienboer, B., Gulcehre, C., Bahdanau, D., Bougares, F.,
  Schwenk, H., Bengio, Y.: Learning phrase representations using rnn
  encoder-decoder for statistical machine translation. In: Empirical Methods in
  Natural Language Processing (2014)

\bibitem{deng2009imagenet}
Deng, J., Dong, W., Socher, R., Li, L.J., Li, K., Fei-Fei, L.: Imagenet: A
  large-scale hierarchical image database. In: Computer Vision and Pattern
  Recognition (2009)

\bibitem{fragkiadaki2015recurrent}
Fragkiadaki, K., Levine, S., Felsen, P., Malik, J.: Recurrent network models
  for human dynamics. In: International Conference on Computer Vision (2015)

\bibitem{goodfellow2014generative}
Goodfellow, I., Pouget-Abadie, J., Mirza, M., Xu, B., Warde-Farley, D., Ozair,
  S., Courville, A., Bengio, Y.: Generative adversarial nets. In: Neural
  Information Processing Systems (2014)

\bibitem{gui2018adversarial}
Gui, L.Y., Wang, Y.X., Liang, X., Moura, J.M.: Adversarial geometry-aware human
  motion prediction. In: European Conference on Computer Vision (2018)

\bibitem{gulrajani2017improved}
Gulrajani, I., Ahmed, F., Arjovsky, M., Dumoulin, V., Courville, A.C.: Improved
  training of wasserstein gans. In: Neural Information Processing Systems
  (2017)

\bibitem{hochreiter1997long}
Hochreiter, S., Schmidhuber, J.: Long short-term memory. Neural computation
  \textbf{9}(8),  1735--1780 (1997)

\bibitem{h36m_pami}
Ionescu, C., Papava, D., Olaru, V., Sminchisescu, C.: Human3.6m: Large scale
  datasets and predictive methods for 3d human sensing in natural environments.
  Transactions on Pattern Analysis and Machine Intelligence  \textbf{36}(7),
  1325--1339 (2014)

\bibitem{kalchbrenner2016neural}
Kalchbrenner, N., Espeholt, L., Simonyan, K., van~den Oord, A., Graves, A.,
  Kavukcuoglu, K.: Neural machine translation in linear time. arXiv preprint
  arXiv:1610.10099  (2016)

\bibitem{karras2019style}
Karras, T., Laine, S., Aila, T.: A style-based generator architecture for
  generative adversarial networks. In: Computer Vision and Pattern Recognition
  (2019)

\bibitem{kingma2013auto}
Kingma, D.P., Welling, M.: Auto-encoding variational bayes. In: International
  Conference on Learning Representations (2014)

\bibitem{ledig2017photo}
Ledig, C., Theis, L., Husz{\'a}r, F., Caballero, J., Cunningham, A., Acosta,
  A., Aitken, A., Tejani, A., Totz, J., Wang, Z., et~al.: Photo-realistic
  single image super-resolution using a generative adversarial network. In:
  Computer Vision and Pattern Recognition (2017)

\bibitem{lee2018stochastic}
Lee, A.X., Zhang, R., Ebert, F., Abbeel, P., Finn, C., Levine, S.: Stochastic
  adversarial video prediction. arXiv preprint arXiv:1804.01523  (2018)

\bibitem{lin2014microsoft}
Lin, T.Y., Maire, M., Belongie, S., Hays, J., Perona, P., Ramanan, D.,
  Doll{\'a}r, P., Zitnick, C.L.: Microsoft coco: Common objects in context. In:
  European Conference on Computer Vision (2014)

\bibitem{martinez2017human}
Martinez, J., Black, M.J., Romero, J.: On human motion prediction using
  recurrent neural networks. In: Computer Vision and Pattern Recognition (2017)

\bibitem{mathieu2015deep}
Mathieu, M., Couprie, C., LeCun, Y.: Deep multi-scale video prediction beyond
  mean square error. In: International Conference on Learning Representations
  (2016)

\bibitem{mirza2014conditional}
Mirza, M., Osindero, S.: Conditional generative adversarial nets. arXiv
  preprint arXiv:1411.1784  (2014)

\bibitem{newell2016stacked}
Newell, A., Yang, K., Deng, J.: Stacked hourglass networks for human pose
  estimation. In: European Conference on Computer Vision (2016)

\bibitem{ohnishi2018hierarchical}
Ohnishi, K., Yamamoto, S., Ushiku, Y., Harada, T.: Hierarchical video
  generation from orthogonal information: Optical flow and texture. In:
  Proceedings of the Thirty-Second {AAAI} Conference on Artificial Intelligence
  (2018)

\bibitem{radford2015unsupervised}
Radford, A., Metz, L., Chintala, S.: Unsupervised representation learning with
  deep convolutional generative adversarial networks. In: International
  Conference on Learning Representations (2016)

\bibitem{ronneberger2015unet}
Ronneberger, O., Fischer, P., Brox, T.: U-net: Convolutional networks for
  biomedical image segmentation. In: Medical Image Computing and
  Computer-Assisted Intervention (2015)

\bibitem{schroff2015facenet}
Schroff, F., Kalenichenko, D., Philbin, J.: Facenet: A unified embedding for
  face recognition and clustering. In: Computer Vision and Pattern Recognition
  (2015)

\bibitem{taylor2007modeling}
Taylor, G.W., Hinton, G.E., Roweis, S.T.: Modeling human motion using binary
  latent variables. In: Neural Information Processing Systems (2007)

\bibitem{vaswani2015attention}
Vaswani, A., Shazeer, N., Parmar, N., Uszkoreit, J., Jones, L., Gomez, A.N.,
  Kaiser, L., Polosukhin, I.: Attention is all you need. In: Neural Information
  Processing Systems (2015)

\bibitem{villegas2017learning}
Villegas, R., Yang, J., Zou, Y., Sohn, S., Lin, X., Lee, H.: Learning to
  generate long-term future via hierarchical prediction. In: International
  Conference on Machine Learning (2017)

\bibitem{vondrick2016generating}
Vondrick, C., Pirsiavash, H., Torralba, A.: Generating videos with scene
  dynamics. In: Neural Information Processing Systems (2016)

\bibitem{walker2017pose}
Walker, J., Marino, K., Gupta, A., Hebert, M.: The pose knows: Video
  forecasting by generating pose futures. In: International Conference on
  Computer Vision (2017)

\bibitem{wang2008gaussian}
Wang, J.M., Fleet, D.J., Hertzmann, A.: Gaussian process dynamical models for
  human motion. Transactions on Pattern Analysis and Machine Intelligence
  \textbf{30}(2),  283--298 (2008)

\bibitem{yan2017skeleton}
Yan, Y., Xu, J., Ni, B., Zhang, W., Yang, X.: Skeleton-aided articulated motion
  generation. In: ACM International Conference on Multimedia (2017)

\bibitem{zhu2017toward}
Zhu, J.Y., Zhang, R., Pathak, D., Darrell, T., Efros, A.A., Wang, O.,
  Shechtman, E.: Toward multimodal image-to-image translation. In: Neural
  Information Processing Systems (2017)

\end{thebibliography}
